\def\year{2022}\relax
\documentclass[letterpaper]{article} 
\usepackage{aaai22}  
\usepackage{times}  
\usepackage{helvet}  
\usepackage{courier}  
\usepackage[hyphens]{url}  
\usepackage{graphicx} 
\def\UrlFont{\rm}  
\usepackage{natbib}  
\usepackage{caption} 
\DeclareCaptionStyle{ruled}{labelfont=normalfont,labelsep=colon,strut=off} 
\usepackage{algorithm}
\usepackage{algorithmic}

\usepackage{graphicx}
\usepackage{amsmath}
\usepackage{booktabs}
\usepackage{multirow}
\usepackage{amssymb}
\usepackage{pifont}
\newcommand{\xmark}{\ding{55}}%
\usepackage{comment}
\usepackage{newfloat}
\usepackage{listings}
\floatstyle{ruled}
\newfloat{listing}{tb}{lst}{}
\floatname{listing}{Listing}
\title{Anatomizing Bias in Facial Analysis}
\author {
    Richa Singh,\textsuperscript{\rm 1}
    Puspita Majumdar, \textsuperscript{\rm 1,2}
    Surbhi Mittal, \textsuperscript{\rm 1}
    Mayank Vatsa \textsuperscript{\rm 1}
}
\affiliations {
    \textsuperscript{\rm 1} IIT Jodhpur \quad
    \textsuperscript{\rm 2} IIIT-Delhi\\
    richa@iitj.ac.in, pushpitam@iiitd.ac.in, mittal.5@iitj.ac.in, mvatsa@iitj.ac.in
}
\usepackage{bibentry}

\begin{document}

\maketitle

\begin{abstract}
Existing facial analysis systems have been shown to yield \textit{biased} results against certain demographic subgroups. Due to its impact on society, it has become imperative to ensure that these systems do not discriminate based on gender, identity, or skin tone of individuals. This has led to research in the identification and mitigation of bias in AI systems. In this paper, we encapsulate bias detection/estimation and mitigation algorithms for facial analysis. Our main contributions include a systematic review of algorithms proposed for understanding bias, along with a taxonomy and extensive overview of existing bias mitigation algorithms. We also discuss open challenges in the field of biased facial analysis.
\end{abstract}

\section{Introduction}
\begin{figure}[t]
\centering
\includegraphics[scale = 0.40]{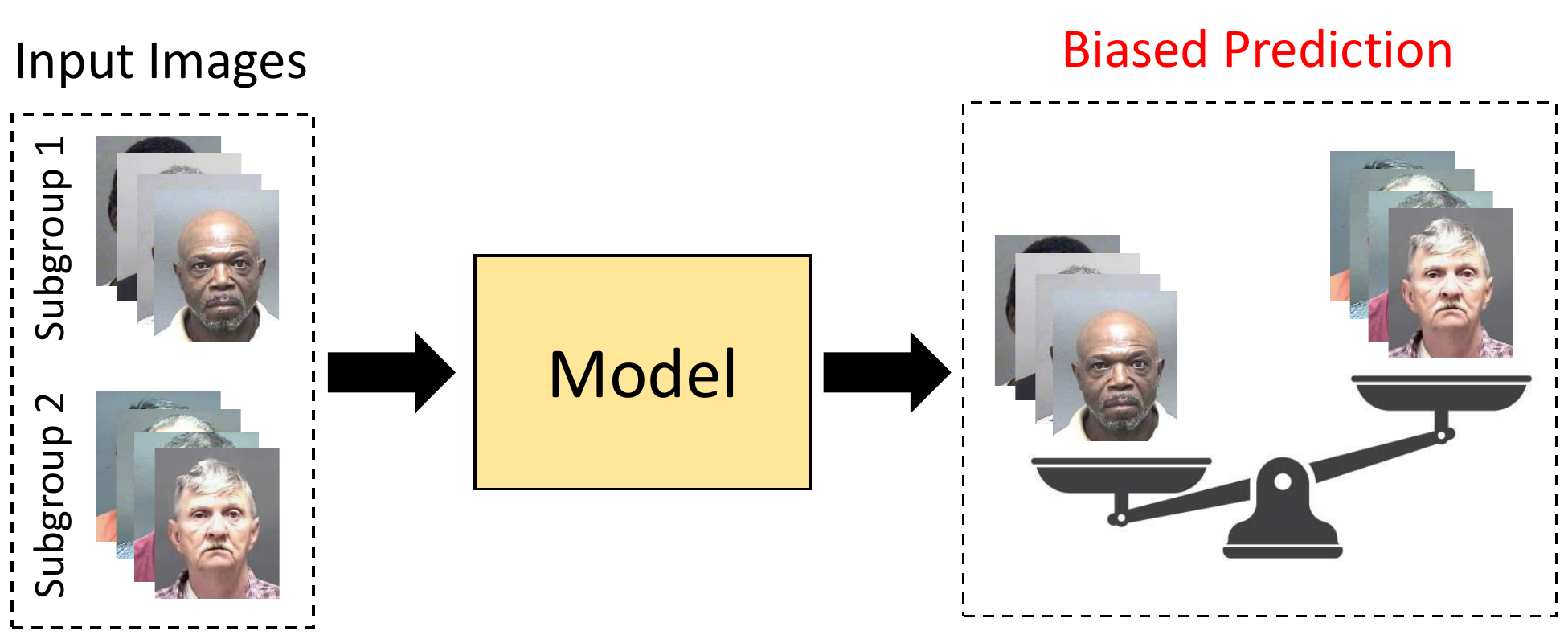}
\caption{Illustrating biased predictions of deep models to favor/disfavor certain demographic subgroups.}
\label{fig:Abstract}
\end{figure}

Artificial Intelligence (AI) systems are used in complex decision-making tasks, including risk assessment, criminal sentencing, and healthcare diagnostics. These decisions affect every aspect of our life, especially when AI systems are used for making predictions about individuals. However, as shown in Figure \ref{fig:Abstract}, several of these systems are found to be biased against certain demographic groups which raises towards trustability and fairness of these systems \cite{osoba2017intelligence}. Multiple instances of biased predictions are observed in AI applications, such as facial analysis tasks, including face detection, attribute prediction, and face recognition. \cite{buolamwini2018gender} have shown the biased behavior of commercial gender classifiers against darker-skinned females. Other instances, such as false identification of members of US Congress as criminals \cite{FRBias} and misclassification of an African American couple as \textit{gorilla} \cite{GooglePhoto}, highlight the biased behavior of existing systems. Due to several of these instances, some corporate and government organizations have banned the use of facial analysis systems \cite{San}.


 These instances have motivated the research community towards designing mechanisms to improve the fairness and trustability of these systems. The focus of researchers is towards understanding bias in model prediction and mitigating its effect to obtain unbiased outcomes \cite{ntoutsi2020bias}. In this regard, multiple analyses have been performed to detect the sources of bias \cite{celis2019learning,krishnapriya2020issues}, databases are proposed to understand the effect of bias \cite{wang2019racial,karkkainen2021fairface}, and algorithms are developed to mitigate the effect of bias in model predictions \cite{majumdar2020subgroup,joo2020gender}.

This research presents a systematic survey of bias in facial analysis tasks as opposed to the broader topic of biometrics addressed in \cite{drozdowski2020demographic}. We provide a synopsis of the analysis towards understanding bias, discuss the bias mitigation algorithms proposed in the literature, and summarize the details of the databases developed for studying bias. As a part of this research, we also present a taxonomy to systematically categorize the techniques proposed for mitigating bias in facial analysis tasks. In the end, with a help of meta-analysis, we discuss open challenges that require attention of the research community. 

\section{Understanding Bias in Facial Analysis}
The prevalence of bias has adverse effects on modern technology, and various attempts have been made to understand and detect the presence of bias. Since bias can occur in a system from various sources (Figure \ref{fig:Bias_FASystems}), different research efforts involve understanding the effect of bias from different perspectives. A large body of work is dedicated towards analyzing biased predictions of models for protected attributes such as gender and ethnic subgroups. In the following subsections, we discuss the research towards understanding bias in facial analysis.  

\subsection{Bias in Face Detection and Recognition}
The performance of face recognition systems has been observed to be inconsistent across different demographic groups. An early observation in this regard is made by \cite{klare2012face} where they demonstrated the difference in recognition performance for different commercial and non-trainable algorithms. They observed a consistently low performance for darker-skinned individuals and prompted the usage of either balanced datasets for training algorithms or separate algorithms for different subgroups. These observations are made in the pre-deep learning era for face recognition algorithms using LBP and Gabor filters. With the onset of the deep learning era, it is observed that demographic bias is an ongoing problem. Research has shown that females tend to have a higher false match rate and a higher false non-match rate as compared to males in face verification applications. To study this phenomenon, the score distributions obtained for genuine as well as impostor pairs across different demographic subgroups have been analyzed. In \cite{albiero2020analysis}, the authors showcase how females with impostor distribution have higher similarity scores while females with genuine distribution have lower similarity scores. In \cite{robinson2020face}, the authors analyzed the decision threshold for the face verification task and observed different thresholds to be optimal for different demographic subgroups. They highlighted how learning a global threshold for matching leads to the incorporation of bias in the system. Similarly, for race, \cite{vangara2019characterizing} used the MORPH dataset and observed that the genuine and impostor distributions are significantly different across Caucasian and African-American subgroups. In another work focusing on bias in face recognition, \cite{krishnapriya2020issues} made similar observations for face verification decision thresholds. They further explore optimal decision thresholds for one-to-many identification search.


\begin{figure}[t]
\centering
\includegraphics[scale = 0.40]{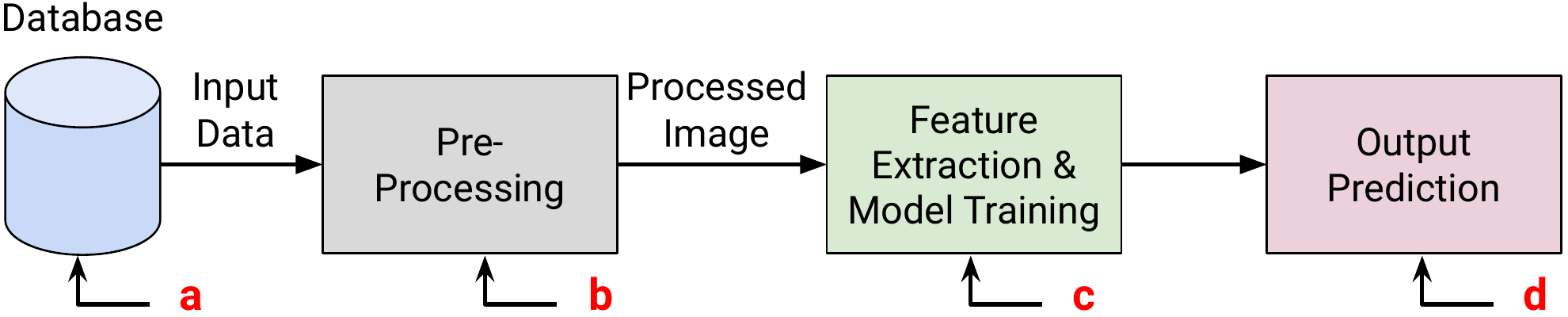}
\caption{Sources of bias in a facial analytics system pipeline. (a) Dataset bias, (b) bias in the pre-processing step, (c) bias in feature extraction and model training, and (d) bias in prediction.}
\label{fig:Bias_FASystems}
\end{figure}

The presence of gender and ethnic subgroup information in face recognition technology clearly indicates that deep models embed the aforementioned information and utilize it for predictions. In this spirit, \cite{acien2018measuring} attempted to infer gender and ethnic group information from pre-trained deep models. They observed that these models classify gender and ethnicity with nearly 95\% accuracy on the LFW database. To further understand how deep models incorporate demographic information, \cite{serna2019algorithmic} used feature space visualizations along with class activation maps (CAMs) which forms a popular technique for inspecting relevant pixels in the input image. They further comment upon how over-representation of certain demographic groups in popular face databases (dataset bias Figure \ref{fig:Bias_FASystems}(a)) has led to popular pre-trained deep face models being biased. Many face recognition systems have a mandatory face quality assessment step while enrolment of an individual. The assessment step ensures the face image meets a certain quality threshold thereby providing a high-quality image for comparison at query time. \cite{terhorst2020face} study the correlation between face quality estimation and demographic bias in face recognition (bias in pre-processing step Figure \ref{fig:Bias_FASystems}(b)). On the evaluation of four algorithms for face quality assessment towards biases to pose, ethnicity, and age, they observed bias towards frontal poses against Asian and African-American ethnicities and towards face images of individuals below 7 years. 

Most face recognition databases collected in the wild lack annotation information for protected attributes such as race and gender. This leads to incomplete information about a model's ability to generalize across different subgroups (dataset bias Figure \ref{fig:Bias_FASystems}(a)). In \cite{kortylewski2019analyzing}, the authors used synthetically generated images for their study and observed significant influence of pose variation on the generalization performance. The authors leveraged synthetic data for analysis and showcased how facial pose and facial identity cannot be completely disentangled by deep networks (bias in model training Figure \ref{fig:Bias_FASystems}(c)). To further study the impact of dataset bias, \cite{gwilliam2021rethinking} analyzed facial recognition performance by training on various imbalanced distributions across race. They observed less biased model predictions after training on a specific subgroup than training on a balanced distribution. Further, the addition of more samples for existing identities in the database improved performance across racial subgroups.

\cite{celis2019learning} analyzed the latent representations of faces to understand potential sources of bias. They observed that the images became brighter with increasing latent values and darker as values got lower, highlighting the importance of skin color in latent representations. To understand where bias is encoded in face recognition, certain works used CAMs and showed how activated regions vary across different demographic subgroups. In such an attempt to understand the cause of bias in deep models, \cite{nagpal2019deep} observed the presence of own-race and own-age bias in popular deep networks. They observed that deep models have a tendency to focus on selected facial regions for a particular ethnic subgroup, with these regions varying across different subgroups. Similarly, in \cite{majumdar2021unravelling}, the authors study the incorporation of bias in model predictions in the presence of real-world distortions (Figure \ref{fig:Bias_FASystems}(d)).


\subsection{Bias in Attribute Prediction}
It has been observed in the literature that many deep learning-based systems are biased in their predictions when measured across different subgroups. \cite{buolamwini2018gender} evaluated the performance of three commercial gender-classification systems across faces with different skin tones. They observed a huge disparity in classification error rates for darker females versus lighter males. \cite{deuschel2020uncovering} studied the impact of gender and skin tone on facial expression detection and used classification accuracy and heatmaps for quantitative and qualitative evaluations. \cite{krishnan2020understanding} investigated the impact of different deep models and training set imbalance on gender classification across different gender-race groups. They highlighted how training set imbalance widens the gap in performance accuracy. The authors further studied facial morphology for different ethnic subgroups using facial landmark detection and obtained interesting insights about probable causes of disparity. To further investigate the impact of different skin tones on gender classification, \cite{muthukumar2019color} used luminance mode-shift and optimal transport techniques to vary the skin tones. They reported that skin tone alone is not the driving factor for observed bias, and broader differences in ethnicity must be considered. \cite{joo2020gender} proposed another approach for understanding bias involves using counterfactuals where they synthesized counterfactual face images with varying gender and ethnic groups keeping the other signals constant. Using these samples, they analyzed performance on different downstream tasks and commented on different hidden biases in the system. Similarly, \cite{denton2019detecting} performed sensitivity analysis based on the performance of deep models using generated counterfactuals. \cite{quadrianto2019discovering} translated the data from the input domain to a fair target domain. They observed interesting outcomes where the model adjusts eyes and lips regions in males to enforce fairness in predictions. Further studying the interdependence between factors leading to biased model predictions, \cite{barlas2021see} analyzed the correlation between context and gender in five proprietary image tagging algorithms. 

The fairness of models is generally attributed to the difference in performance across subgroups. In a different direction, \cite{serna2021insidebias} proposed \textit{InsideBias} in which they studied how the model represents the information instead of how it performs. They analyzed the features learned by the models while training using an unbalanced dataset in the context of bias. \cite{serna2021ifbid} showed how analyzing model weights provides insights into its biased behavior. \cite{li2021discover} have proposed a variation loss that optimizes the hyperplane in the latent space to obtain biased attribute information. Researchers continue to study the influence of bias in predictions (Figure \ref{fig:Bias_FASystems}(d)), and predominantly how it might have been incorporated at the data and algorithm level (Figures \ref{fig:Bias_FASystems}(a) and \ref{fig:Bias_FASystems}(c)). A large body of work focuses on analyzing popular deep model architectures and COTS algorithms. Popular tools include feature map visualizations, fairness and performance evaluation metrics across subgroups, generated counterfactuals, and skewed training. However, a limited number of studies have been performed to detect bias-inducing factors in an automated manner and require the attention of the research community.

\section{Bias Mitigation in Facial Analysis}
In this section, we provide a taxonomy of the bias mitigation techniques proposed for various face analysis tasks. 

\subsection{Face Detection and Recognition}
Majority of the algorithms developed for mitigating bias in face detection and recognition are based on \textbf{deep learning} based approaches. These approaches include designing novel loss functions, custom networks, and discrimination-aware learning methods. \cite{amini2019uncovering} proposed a novel algorithm for mitigating hidden bias in face detection algorithms. The proposed algorithm uses a variational autoencoder to learn the latent structure within the database. The learned latent distributions are used to re-weight the importance of certain data points during training. In the literature, it is shown that imbalanced class distribution leads to biased predictions of deep models. To handle the problem of imbalanced class distribution, \cite{huang2019deep} proposed to learn Cluster-based Large Margin Local Embedding, and combined it with k-nearest cluster algorithm for improved recognition performance. A deep information maximization adaptation network is proposed by \cite{wang2019racial} for bias mitigation using deep unsupervised domain adaptation techniques. The authors considered Caucasian as the source domain and other races as target domains to decrease race gap at domain-level. \cite{terhorst2020comparison} proposed a fairness driven neural network classifier that works on the comparison-level of a biometric system. A novel penalization term in the loss function is used to train the proposed classifier to reduce the intra-ethnic performance differences and introduce both group and individual fairness to the decision process. \cite{bruveris2020reducing} proposed to mitigate bias in face recognition due to imbalanced data distribution by employing sampling strategies that balance the training procedure. Attention mechanisms are also used for enhancing the fairness of recognition models. In this direction \cite{gong2021mitigating} proposed group adaptive classifier that uses adaptive convolution kernels and attention mechanism for bias mitigation. The adaptive module and the attention maps help to activate different facial regions to learn discriminative features for each demographic group. \cite{yang2021ramface} proposed the race adaptive margin based face recognition (RamFace) model to enhance the discriminability of the features. The authors also proposed a race adaptive margin loss function to automatically derive different optimal margins to mitigate the effect of racial bias. 


Some researchers have used \textbf{adversarial learning} based approaches for bias mitigation. \cite{alasadi2019toward} presented a framework for matching low resolution and high-resolution facial images. The aim is to mitigate bias in cross-domain face recognition. The proposed framework consists of two parts, the first part maximizes the face matching quality and the second part minimizes the prediction of demographic properties to reduce bias in model prediction. A novel de-biasing adversarial network is proposed by \cite{gong2020jointly} that adversarially learns to generate disentangled representations for unbiased face and demographics recognition. The proposed network consists of four classifiers to distinguish the identity, gender, race, and age of the facial images. \cite{dhar2020adversarial} presented a novel Adversarial Gender De-biasing algorithm to reduce the gender prediction ability of face descriptors. The proposed algorithm unlearns the gender information in descriptors while training them for classification.

Researchers have also proposed techniques that could be combined with \textbf{pre-trained} models to improve their performance and reduce bias in model prediction. \cite{serna2020sensitiveloss} proposed a discrimination-aware learning method, termed as Sensitive loss for bias mitigation. The proposed loss function is based on the triplet loss function and a sensitive triplet generator to improve the performance of pre-trained models. A novel unsupervised fair score normalization approach based on individual fairness is proposed by \cite{terhorst2020post}. The proposed solution is easily integrable into existing systems that reduce bias and improves the overall recognition performance. 

\begin{table*}[]
\centering
\caption{Details of the databases used for bias study.}
\renewcommand{\arraystretch}{1.1}
\begin{tabular}{|l|c|c|c|c|c|c|}
\hline
\textbf{Database}                                                                                                         & \textbf{Identity} & \textbf{Gender} & \textbf{Race} & \textbf{Age} & \textbf{No. of Subjects}                                & \textbf{No. of Images}                                    \\ \hline
MORPH \cite{rawls2009morph}                                                                                               & \checkmark        & \checkmark      & \checkmark    & \checkmark   & 13,000+                                                    & 55,000+                                                      \\ \hline
IMFDB \cite{setty2013indian}                                                                                              & \checkmark        & \checkmark      & \xmark        & \checkmark   & 100                                                     & 34,512                                                    \\ \hline
Adience \cite{eidinger2014age}                                                                                            & \checkmark        & \checkmark      & \xmark        & \checkmark   & 2,284                                                   & 26,580                                                    \\ \hline
CACD \cite{chen2015face}                                                                                                  & \checkmark        & \xmark          & \xmark        & \checkmark   & 2,000                                                   & 163,446                                                   \\ \hline
LFWA \cite{huang2008labeled}                                                                                              & \checkmark        & \checkmark      & \xmark        & \xmark       & 5,749                                                   & 13,233                                                    \\ \hline
CelebA \cite{liu2015faceattributes}                                                                                       & \xmark            & \checkmark      & \xmark        & \xmark       & 10,000+                                                    & 2,02,599                                                  \\ \hline
AgeDb \cite{moschoglou2017agedb}                                                                                          & \checkmark        & \xmark          & \xmark        & \checkmark   & 568                                                     & 16,488                                                    \\ \hline
AAF \cite{cheng2019exploiting}                                                                                            & \xmark            & \checkmark      & \xmark        & \checkmark   & -                                                       & 13,298                                                    \\ \hline
IJB-C \cite{maze2018iarpa}                                                                                                & \checkmark        & \checkmark      & \xmark        & \xmark       & 3500                                                    & 33,000                                                       \\ \hline
UTKFace \cite{zhang2017age}                                                                                               & \xmark            & \checkmark      & \checkmark    & \checkmark   & -                                                       & 20,000+                                                      \\ \hline
RFW \cite{wang2019racial}                                                                                                 & \checkmark        & \xmark          & \checkmark    & \xmark       & 40,607                                                  & 11,430                                                    \\ \hline
\begin{tabular}[c]{@{}l@{}}BUPT-Balancedface, BUPT-Globalface,\\ BUPT-Transferface \cite{wang2020mitigating}\end{tabular} & \checkmark        & \xmark          & \checkmark    & \xmark       & \begin{tabular}[c]{@{}c@{}}28,000, 38,000,\\ 10,000\end{tabular} & \begin{tabular}[c]{@{}c@{}}1.3M, 2M,\\ 0.6M+\end{tabular} \\ \hline
DiveFace \cite{morales2020sensitivenets}                                                                                  & \checkmark        & \checkmark      & \checkmark    & \xmark       & 24,000                                                     & 72,000                                                       \\ \hline
FairFace \cite{karkkainen2021fairface}                                                                                    & \xmark            & \checkmark      & \checkmark    & \checkmark   & -                                                       & 108,501                                                   \\ \hline
\end{tabular}
\label{tab:db}
\end{table*}


\textbf{Generative} approaches are adopted by researchers to synthesize images of the under-represented class for bias mitigation. \cite{mcduff2019characterizing} proposed a simulation-based approach using generative adversarial models for synthesizing facial images to mitigate gender and racial biases in commercial systems. A novel data augmentation methodology is proposed by \cite{yucer2020exploring} to balance the training database at a per-subject level. The authors used image-to-image transformation for transferring facial features with sensitive racial characteristics while preserving the identity-related features for mitigating racial bias. Recently, \textbf{reinforcement learning} based approach is used by \cite{wang2020mitigating} to learn balanced features and remove racial bias in face recognition using the idea of adaptive margin.

It is observed that the majority of the algorithms for mitigating bias in face recognition are designed to mitigate bias for a specific demographic group. Therefore, these algorithms may not generalize for other demographic groups \cite{xu2021consistent}. It is our belief that researchers should focus more on designing solutions that are generalizable across different demographic groups.

\subsection{Attribute Prediction}
Several algorithms have been proposed for bias mitigation in attribute prediction. Among \textbf{deep learning} approaches, one set of algorithms aim to unlearn the model's dependency on sensitive attributes. \cite{kim2019learning} proposed a regularization loss to minimize the mutual information between feature embedding and bias to unlearn the bias information. Attribute aware filter drop is proposed by \cite{nagpal2020attribute} that performs the primary attribute classification task while unlearning the dependency of the model on sensitive attributes. \cite{tartaglione2021end} proposed a regularization strategy to disentangle the biased features while entangling the features belonging to the same target class. Apart from this some techniques are proposed that use feature distillation \cite{jung2021fair} and mutual information between the learned representation \cite{ragonesi2021learning}. For understanding different bias mitigation algorithms \cite{wang2020towards} provided a thorough analysis of the existing bias mitigation techniques. They further designed a domain-independent training technique for bias mitigation. An interesting data augmentation strategy, \textit{fair mixup} is proposed by \cite{chuang2021fair} to optimize group fairness constraints. The authors proposed to regularize the model on interpolated distributions between different subgroups of a demographic group. Recently, \cite{park2021learning} argued that removing the information of sensitive attributes in the decision process has the limitation of eliminating beneficial information for target tasks. To overcome this, the authors proposed Fairness-aware Disentangling Variational Auto-Encoder for disentangling data representation into three latent subspaces. A decorrelation loss is proposed to align the overall information into each subspace, instead of removing the information of sensitive attributes.


\cite{dwork2018decoupled} proposed a decoupling technique for bias mitigation of \textbf{black-box} ML algorithms. The proposed technique learns separate classifiers for different groups to increase fairness and accuracy in classification systems. \cite{nagpal2020diversity} proposed diversity blocks to de-bias existing models. The diversity block is trained using small training data and can be added to any black-box model. \cite{roh2020fairbatch} addressed the problem of bias mitigation using bi-level optimization. They proposed to adaptively select mini-batches for improving model fairness. A bias mitigation algorithm based on adversarial perturbation is proposed by \cite{majumdar2020subgroup}. The proposed algorithm learns a subgroup invariant perturbation to be added to the input database to generate a transformed database. The transformed database, when given as input to a model, produces unbiased outcomes. The proposed algorithm is able to mitigate bias in \textbf{pre-trained} model prediction without re-training. Recently, \cite{majumdar2021attention} proposed an algorithm based on attention mechanism for mitigating bias in pre-trained models.


Oversampling the minority class is one of the popular techniques for handling the class imbalance problem. Imbalance in class distribution leads to biased predictions and multiple \textbf{generative} approaches have been proposed for mitigation. \cite{ramaswamy2020fair} used generative adversarial networks (GANs) to generate images for data augmentation. The generated images are perturbed in the latent space to generate balanced training data w.r.t protected attribute for bias mitigation. A weakly-supervised algorithm is proposed by \cite{choi2020fair} to overcome database bias for deep generative models. An additional unlabeled database is required by the proposed approach for bias detection in existing databases. \cite{tan2020improving} proposed an effective method for improving the fairness of image generation for a pre-trained GAN model without retraining. Images generated using the proposed method are applied for bias quantification in commercial face classifiers. A multi-attribute framework is proposed by \cite{georgopoulos2021mitigating} to transfer facial patterns even for the underrepresented subgroups. The proposed method helps to mitigate dataset bias by data augmentation. Apart from the generative approaches, \cite{wang2019balanced} proposed an \textbf{adversarial} approach to remove bias from intermediate representations of a deep neural network. The proposed approach reduces gender bias amplification and maintains the overall model performance. \cite{adeli2021representation} used adversarial training to maximize the discriminative power of the learned features with respect to the main task and minimize the statistical mean dependence with the bias variable. By minimizing the dependency on the bias variable, the authors have shown reduced effect of bias in model prediction.

Majority of the algorithms are focused on alleviating the influence of demographics on model predictions for enhancing fairness. However, \cite{terhorst2020beyond} demonstrated that face templates store non-demographic characteristics as well. Our assertion is that the biased prediction of models could be due to the non-demographic characteristics stored in the face images. Therefore, we believe that considering the non-demographic attributes during bias mitigation is important for designing robust systems.


\section{Facial Analysis Databases for Bias Study}
\label{FA_Db}
Several facial analysis databases have been proposed in the literature. Initially, existing databases with demographic information were used for studying the effect of bias. \cite{rawls2009morph} proposed the MORPH database with subjects belonging to different gender subgroups and ethnicity within the age range of 16 to 77 years. It is one of the largest databases for studying the effect of age on face recognition algorithms. The Indian Movie Face Database (IMFDB) proposed by \cite{setty2013indian} consists of face images of Indian actors and actresses. The database is annotated with age, gender, pose, expression, and the amount of occlusion present in an image. \cite{eidinger2014age} proposed the Adience database containing face images labeled with age and gender information. The database is mainly used for age estimation. The Cross-Age Celebrity Dataset (CACD) proposed by \cite{chen2015face} contains images of celebrities with age ranging from 16 to 62 years. The LFWA \cite{huang2008labeled} and CelebA \cite{liu2015faceattributes} databases contain 40 annotated facial attributes and are commonly used for attribute prediction tasks. These databases are used for recent research that focuses on analyzing the performance of attribute prediction across protected attributes (male and young), followed by developing algorithms for bias mitigation \cite{tan2020improving}. The AgeDb \cite{moschoglou2017agedb} and All-Age Faces (AAF) \cite{cheng2019exploiting} databases are used for analyzing the performance of algorithms across age subgroups. Apart from this, the IJB-C database \cite{maze2018iarpa} is one of the largest databases with skin-tone information on the Fitzpatrick scale used for studying bias. 

With the increased attention on understanding different aspects of bias in model prediction and developing fair algorithms, multiple databases are proposed for studying bias. \cite{zhang2017age} proposed the UTKFace database with more than 20K images having variations in pose, illumination, expression, occlusion, and resolution. \cite{buolamwini2018gender} proposed the Pilot Parliaments Benchmark (PPB) database for studying the effect of bias in gender classifiers w.r.t different skin tones. The PPB database consists of 1270 subjects from three African and three European countries. Authors labeled the skin tones of the subjects using the Fitzpatrick six-point labeling system. Racial Faces in-the-Wild (RFW) database proposed by \cite{wang2019racial} is an unconstrained testing database for studying racial bias in face recognition. The database consists of Caucasian, Indian, Asian, and African racial subgroups. \cite{wang2020mitigating} proposed four different training databases, namely BUPT-Balancedface (race-balanced), BUPT-Globalface (racial distribution approximately equal to the real distribution of world’s population), BUPT-Transferface containing labeled data for Caucasian and unlabeled data for other subgroups (created for unsupervised domain adaptation), and MS1M-wo-RFW containing non-overlapping subjects of MS-Celeb-1M database for studying the effect of race on face recognition algorithms. \cite{morales2020sensitivenets} proposed the DiveFace database with annotations for gender and ethnicity (Caucasian, European, Asian) subgroups. Recently, the FairFace database is proposed \cite{karkkainen2021fairface}, which is balanced across seven race subgroups: White, Black, East Asian, Middle Eastern, Southeast Asian, Indian, and Latino. The databases proposed for studying the effect of bias, shown in Table \ref{tab:db}, have escalated the research towards understanding bias and boosted the development of algorithms for bias mitigation. 

\begin{figure}[t]
\centering
\includegraphics[scale = 0.33]{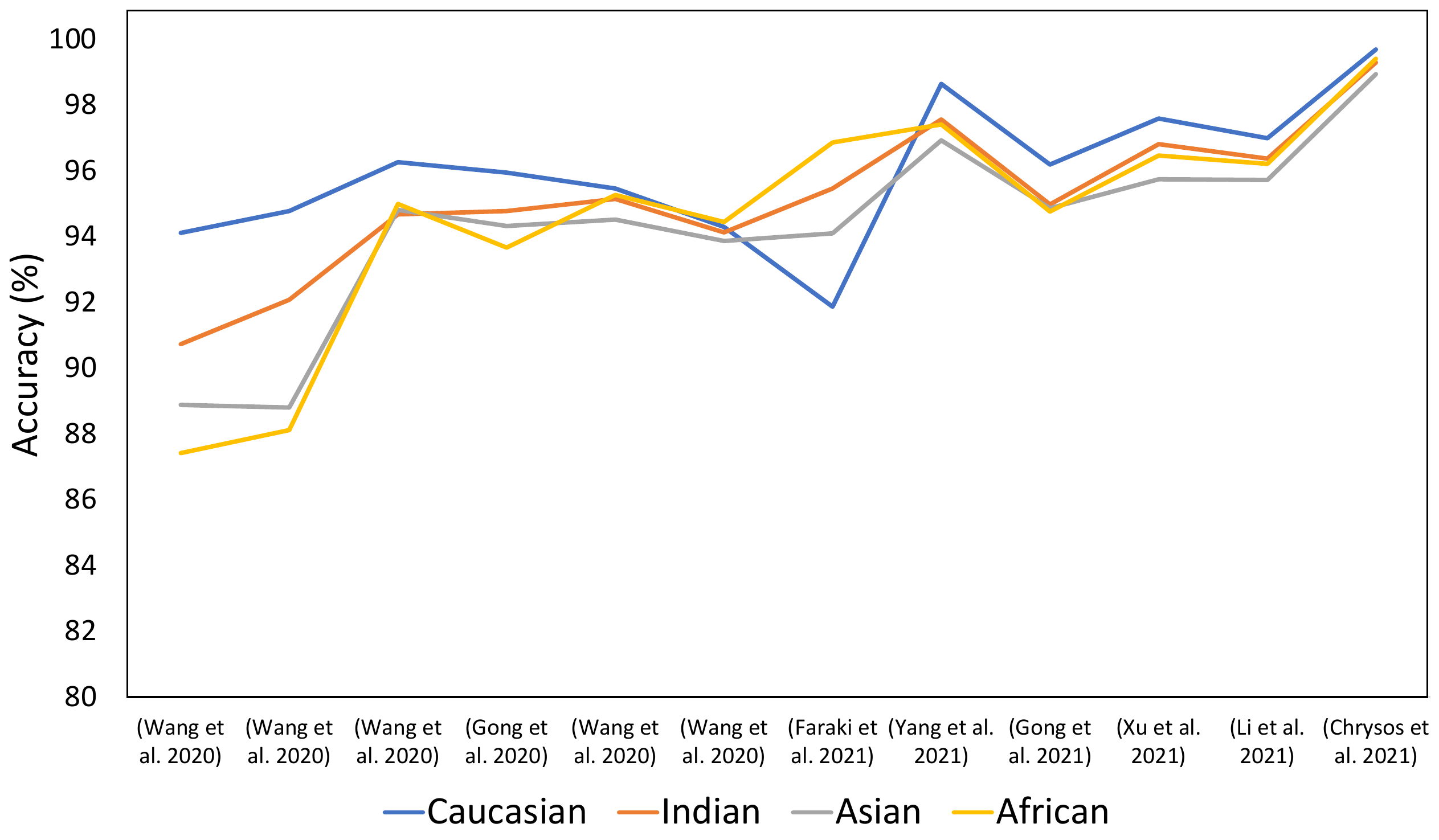}
\caption{Meta-analysis of the verification accuracy reported on the RFW dataset across the four racial subgroups.}
\label{fig:MetaAnalysis}
\end{figure}

\section{Open Challenges}
Research towards bias and fairness has gained significant advancements, and several solutions have been proposed to improve the trustability and dependability of facial analysis systems. The trend in deep learning models has been towards improving fairness for different demographic subgroups. As illustrated in Figure \ref{fig:MetaAnalysis}, newer algorithms have shown improved fairness across different racial subgroups. Despite the progress achieved in designing fair solutions, there are various open challenges that require the attention of the research community. Here, we discuss some of the challenges that require more attention and focused research efforts.

\noindent \textbf{Fairness in Presence of Occlusion:} Wearing face masks has become a mandate in public places worldwide due to the COVID-19 pandemic. Thus, face recognition algorithms are required to recognize faces in the presence of masks. Masks occlude a major portion of the facial region that poses challenges to face recognition algorithms. To facilitate research in this direction, researchers have proposed multiple masked face databases. However, these databases contain limited demographic information. In a real-world scenario, it is important that the face recognition algorithms perform equally well across different demographic groups in the presence of occlusion. Limited research is done towards understanding the effect of bias in the presence of occlusion and more attention is required to develop fair algorithms.  

\noindent \textbf{Fairness Across Intersectional Subgroups:} Majority of the research is performed to mitigate bias due to a single demographic group. Less attention is paid towards bias mitigation across intersectional subgroups. For instance, a model that is fair across gender subgroups may be biased towards old darker-skinned females. To ensure trustability, it is important that the output predictions of a model are fair across individual as well as intersectional demographic subgroups. Thus, more research is required towards identifying and mitigating bias across intersectional subgroups.

\noindent \textbf{Trade-off Between Fairness and Model Performance:} Bias mitigation may affect the overall model performance. While mitigation algorithms increase fairness, and the trained models achieve equal performance across different demographic subgroups, the model performance on over-represented subgroups may reduce. It is challenging to simultaneously reduce the effect of bias without hampering the overall model performance.  

\noindent \textbf{Benchmark Databases:} Multiple databases have been proposed in recent years for studying bias as shown in Table \ref{tab:db}. A very limited set of databases provide identity information along with demographic information. Even for databases that provide both information, there is a clear lack of consistency among the demographic information provided. For example, different databases segregate data for different number of ethnic subgroups. We need to account for intra-class variations within demographic groups such as Indian and Asian which have huge diversity with respect to skin-tone and facial appearance. None of the existing databases provide the distribution of subgroups (based on skin tone or facial appearance) within an ethnicity. This poses challenge regarding interpretability of model predictions. Further, an imbalance in database w.r.t the unlabeled demographic groups may introduce bias in model prediction. In such scenarios, it becomes difficult to interpret the source of bias. Therefore, we believe that construction of large-scale benchmark databases with details of demographic information will help to decipher the cause of bias in model prediction and develop algorithms for bias mitigation.

\noindent \textbf{Unavailability of Complete Information:} Existing algorithms are based on the assumption that demographic information is available during training. However, due to privacy concerns and regulations in the real-world, the collection of demographic information or their use during training is precluded. Further, the protected attribute information can be noisy. This severely limits the applicability of existing bias mitigation algorithms and demands the need for the development of algorithms which do not require demographic information for bias mitigation.

\section{Conclusion}
Fairness in model prediction is important for the trustability of AI systems. In recent years, growing attention is observed towards handling the problem of bias in model predictions, and significant progress is made towards understanding bias and developing mitigation algorithms. We presented a systematic review of bias in facial analysis tasks and provided a taxonomy of the algorithms proposed for bias mitigation. We also discuss some of the open challenges that require proper attention and continuous research efforts. We believe that solving or providing solutions for these open research problems is necessary for further reducing the accuracy gap across different subgroups and building trusted AI systems.  

\section*{Acknowledgements}
P. Majumdar is partly supported by DST Inspire Ph.D. Fellowship. S. Mittal is partially supported by UGC-Net JRF Fellowship. M. Vatsa is partially supported through Swarnajayanti Fellowship. This research is also partially supported by Facebook Ethics in AI award.

\bibliography{aaai22}

\end{document}